\documentclass{bmvc2k}

\usepackage{graphicx}
\usepackage{amsfonts}
\usepackage{caption}
\usepackage{multirow}
\usepackage{bm} 
\makeatletter
\newcommand{\figcaption}[1]{\def\@captype{figure}\caption{#1}}
\newcommand{\tblcaption}[1]{\def\@captype{table}\caption{#1}}
\makeatother

\title{Guidance-base Diffusion Models for Improving Photoacoustic Image Quality}

\addauthor{Tatsuhiro Eguchi}{tatsuhiro.eguchi@human.ait.kyushu-u.ac.jp}{1}
\addauthor{Shunpei Takezaki}{shumpei.takezaki@human.ait.kyushu-u.ac.jp}{1}
\addauthor{Mihoko Shimano}{miho@nii.ac.jp}{2}
\addauthor{Takayuki Yagi}{yagi.takayuki@luxonus.jp}{3}
\addauthor{Ryoma Bise}{bise@ait.kyushu-u.ac.jp}{1}

\addinstitution{
 Kyushu University\\
 Fukuoka, Japan
}
\addinstitution{
 National Institute of Informatics\\
}
\addinstitution{
 Luxonus Inc.\\
}

\runninghead{Student, Prof, Collaborator}{}

\begin{document}

\maketitle

\begin{abstract}
Photoacoustic(PA) imaging is a non-destructive and non-invasive technology for visualizing minute blood vessel structures in the body using ultrasonic sensors. In PA imaging, the image quality of a single-shot image is poor, and it is necessary to improve the image quality by averaging many single-shot images. Therefore, imaging the entire subject requires high imaging costs.
In our study, we propose a method to improve the quality of PA images using diffusion models. In our method, we improve the reverse diffusion process using sensor information of PA imaging and introduce a guidance method using imaging condition information to generate high-quality images.
\end{abstract}

%-------------------------------------------------------------------------
\section{Introduction}
\label{sec:intro}
Photoacoustic(PA) imaging\cite{li2009photoacoustic} is a technique that visualizes the fine vascular structures within the body. 
In PA imaging, blood vessels absorb laser energy from short-pulsed near-infrared light and convert the energy into heat, leading to the emission of ultrasonic waves. The structures of objects can be reconstructed by sensing the emitted photoacoustic waves.
%Ultrasonic signals generated from vessels are acquired by sensors and then converted into images. 
%Ultrasound is generated by illuminating blood vessels with laser light, which the blood absorbs, and the resulting signals are captured by sensors and visualized. 
This technology is non-destructive and non-invasive and is used to understand vascular structures before surgery~\cite{BiseR2016vessel,SaitoS2019,kikkawa2019ssl}.

%The problem with PA imaging is that a single-shot image, which captures a small local area, is low quality, which contains many noise due to limitations of the number of acoustic sensors. The foreground parts (blood vessels) are often partially missing (Figure.~\ref{fig:pa} right-upper).
%To reduce noise, image-averaging techniques are effective, which take the average of many scans at the same position based on the assumption that the vessels are linearly correlated but the noise is random~\cite{biseMiccai2016,asanomiMiccai2022}. 
%It requires the acquisition time to capture the wide area of a body. However, the data acquisition speed of PA imaging is limited by the laser repetition rate; the number of sampling is usually limited for real-time imaging and reducing the patient burden.
 
\begin{figure}[t]
    \centering
    \vspace{1mm}
    \includegraphics[scale=0.65]{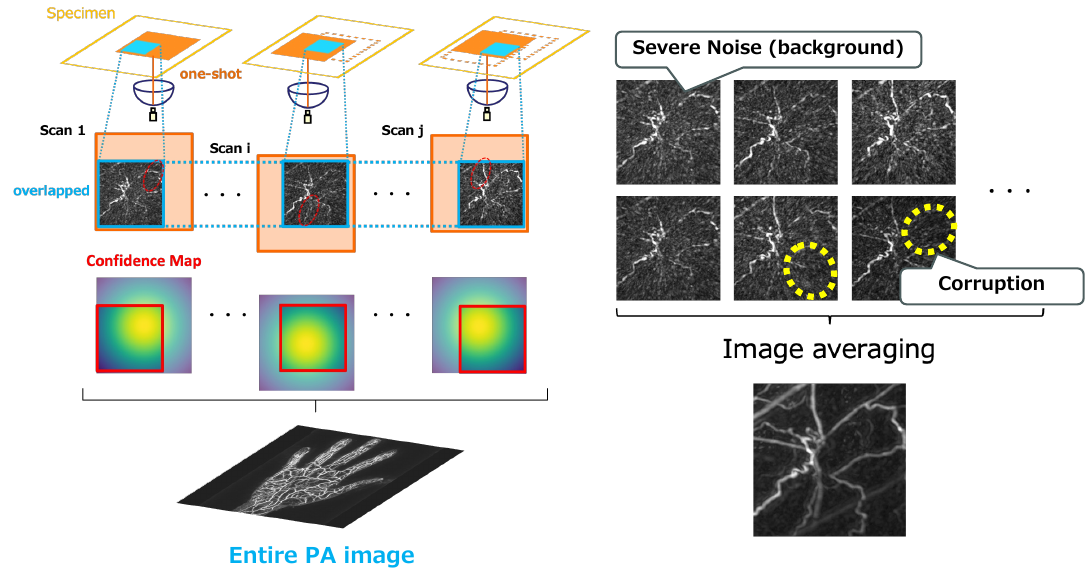}
    \vspace{2mm}
    \caption{{\bf Left}: Overall mechanisim of photoacoustic imaging. Heat map is the confidence map based on the position of light exposure.
{\bf Right}: The upper images are single-shot images (low-quality), and the bottom image is an averaged image (high-quality).}
    \vspace{-3mm}
    \label{fig:pa}
\end{figure}

The issue with PA imaging is that single-shot images, which capture small local areas, are of low quality due to the limitations of the number of acoustic sensors. These images often contain significant noise, leading to a partial absence of foreground parts (such as blood vessels) (Figure~\ref{fig:pa}, upper-right).
Image-averaging techniques effectively reduce noise, which involves averaging multiple scans at the same position under the assumption that vessels are linearly correlated while noise is random~\cite{biseMiccai2016,asanomiMiccai2022}. However, this approach requires increased acquisition time to cover a wide body area.
Moreover, the data acquisition speed of photoacoustic imaging is limited by the laser repetition rate, and the number of samples is typically restricted to enable real-time imaging and minimize patient burden.
This study aims to transform low-quality single-shot images into a high-quality image.

To achieve this goal, we propose a method using diffusion models guided by imaging condition information. Diffusion models, a recently popular image generation model, enables the creation of diverse, high-quality images\cite{dhariwal2021diffusion,graikos2022diffusion,ho2020denoising,saharia2022photorealistic,song2021denoising}. In image-to-image tasks, diffusion models have been widely used and achieved high accuracy \cite{rahman2023ambiguous,saharia2022palette,saharia2022image}.

In this study, we use diffusion models to generate high-quality PA images from single-shot images that include noise and missing foreground elements. Specifically, we utilize the guidance that uses multiple single-shot images (multi-shots) rather than a single-shot image (single-shot), resulting in higher-quality images. We guide the reverse diffusion model towards higher quality by using the vector from the noise estimated from the low-quality single-shot images to the noise estimated from the multi-shot images.

Moreover, this method introduces the unique property of image averaging in PA imaging into the noise estimation of the diffusion model. Specifically, as shown in Figure~\ref{fig:pa}, the laser light scatters inside the body. It spreads with a Gaussian distribution from the irradiation position, weakening the signal strength as it moves away from this position. As shown in Figure~\ref{fig:pa}, the relative positions of light irradiation in single-shot images vary, leading to areas with clear blood vessels and areas with missing details (i.e., regions of differing quality).

Therefore, we propose a method that combines the estimated noise from each single-shot image, considering the reliability of the signals based on the light irradiation positions. This allows for the estimation of high-quality images. We conducted experiments using actual PA images and confirmed the effectiveness of our method compared to traditional techniques.

% \vspace{-mm}
\section{Related Work}
\label{sec:relatedwork}
%\subsection{Supervised Denoising Methods}
\noindent\textbf{Supervised Denoising Methods.} 
Numerous methods for image noise reduction have been developed to date. Research by Zhang~\cite{zhang2017beyond}, Zamir~\cite{zamir2022restormer}, and others have proposed methods using CNNs and Vision Transformers. Furthermore, Luo et al.~\cite{pmlr-v202-luo23b} have achieved noise removal using diffusion models. Recently, diffusion models have been widely used for medical images. Particularly in denoising tasks, Gao et al.~\cite{corediff10268250} proposed a method applying diffusion models for noise reduction in low-dose CT images. Methods using diffusion models for noise removal have also been proposed for positron emission tomography images and ultrasound images~\cite{ultra10.1007/978-3-031-53767-7_19, pet10.1007/978-3-031-43907-0_26}. All these methods focus on general images containing synthetic Gaussian noise, and to our knowledge, no studies have specifically targeted the quality improvement of photoacoustic images. In addition, no method uses the characteristic of image averaging for noise reduction.
Therefore, our approach, which effectively utilizes imaging condition information, is considered superior in improving the quality of images.

%\subsection{Guidance Methods for Diffusion Models}
\vspace{0.5\baselineskip}
\noindent\textbf{Guidance Methods for Diffusion Models.} 
% 2024/05/09 09:02 s.takezaki
% 
As a method for conditional image generation based on specific classes using diffusion models, a Classifier-Guidance technique utilizes the classifier's gradients with the estimates of diffusion models\cite{dhariwal2021diffusion,song2021scorebased}, enabling more stable image generation that reflects class information. 
In Classifier-free Guidance\cite{ho2021classifierfree}, vectors of the conditional and unconditional diffusion models are used instead of using classifier gradients.
%based on the following equation:
%\begin{equation}
%    \tilde{\bm{\epsilon}}_\theta(\bm{x}_t,t,c) = (1+w)\bm{\epsilon}_\theta(\bm{x}_t,t,c) - w\bm{\epsilon}_\theta(\bm{x}_t,t,\phi),
%  \label{eq:cfg}
%\end{equation}
%where $w$ represents the strength of guidance, and $\phi$ signifies the unconditional token used for unconditional generation. Adjusting $w$ achieves a trade-off between image generation diversity and class information fidelity. 
This method is not limited to class-conditional generation and has also been used in Text-to-Image tasks based on textual information \cite{gafni2022make,Ramesh:2022esg,rombach2022high}. Additionally, there exists guidance using a CLIP model and methods by guiding the internal representations of diffusion models, which are applicable to tasks such as segmentation and object detection\cite{s2021glide,bansal2024universal,10484388,hu2023selfguided}.
To the best of our knowledge, no studies have applied such guidance in denoising tasks. In this paper, we introduce a guidance mechanism that utilizes vectors from lower-quality to higher-quality images to enhance the generation of high-quality images.

\section{Prepare a paired dataset (low and high-quality images)}

Our method aims to train a diffusion model-based image transfer from a low-quality to a high-quality image. To achieve this, pairs of low-and-high-quality images are required for supervised training data. In this section, we explain how to prepare the dataset.

%In PA imaging, multi-scanning is necessary to cover a large area of the human body, as illustrated in Figure.~\ref{fig:pa}. 
%The data acquisition speed of photoacoustic imaging is constrained by the laser repetition rate and the number of ultrasound sensors; typically, the number of scans is limited to enable real-time imaging and reduce system costs. 
%One approach to reducing the number of scans is by employing a sliding window scanning technique. However, a single-shot image reconstructed by PA imaging often suffers from significant noise due to sound and light scattering and limitations in sensor layout. Consequently, the overall image quality is compromised.

We employed an image-averaging technique with image alignment based on~\cite{biseMiccai2016} to produce high-quality images. This method exploits the fact that the average of random noise in the background becomes a small constant while the linearly correlated foreground becomes more prominent. 
To apply the image-averaging technique, single-shot images taken while moving the light source and sensor positions are averaged over overlapping imaging areas (see Figure~\ref{fig:pa} left).
%In multiple-scanning, the location of short-pulsed near-infrared laser irradiation is slightly different, even though capturing the same location as shown in Figure~\ref{fig:pa}.
Image averaging proves effective when the sample is static. 
To address issues arising from patient movement during scans, images are aligned using \cite{biseMiccai2016}. This method can produce high-quality images as the number of scans is increased.

For our training data, we captured specimens using many scans to produce high-quality images, and we decreased the number of scans to generate low-quality images.
In this paper, we used an image generated by $M$ images in the same location as a low-quality image and that generated by 20 to 40 images as a high-quality one. 
The paired images of low-and-high-quality images are denoted as $\{\{L_{1}^i, \ldots, L_{M}^i\}, H^i\}_{i=1}^N$. $L^i$ indicates a single-shot image in the same position $i$, $M$ is a small number, and $H^i$ indicates the corresponding high-quality image in the $i$-th position, which is generated using many scans.

Our method utilizes the distribution map of scattered light in the skin, corresponding to $L^i_m$. This involves irradiating light into the body, which then scatters as it spreads throughout the body. As the absorbed light is large, the magnitude of acoustic signals becomes larger.

As described above, the laser irradiation locations of these $M$ single-shot images $L_{1}^i, \ldots, L_{M}^i$ are different, where the center of the location is recorded in the imaging system as shown in Figure~\ref{fig:pa}. We model the scattering light distribution as the Gaussian distribution, denoted by $\bm{h}^i_m$ for $m$-th scan at location $i$.

Note that generating high-quality images through multiple scans with image alignment is unsuitable for real-world applications because performing multiple scans and alignment for all images takes a long time, increasing the burden on patients. Therefore, we propose an image transfer method from low to high quality.
\begin{figure*}[t]
    \centering
    \includegraphics[scale=0.3, width=\linewidth]{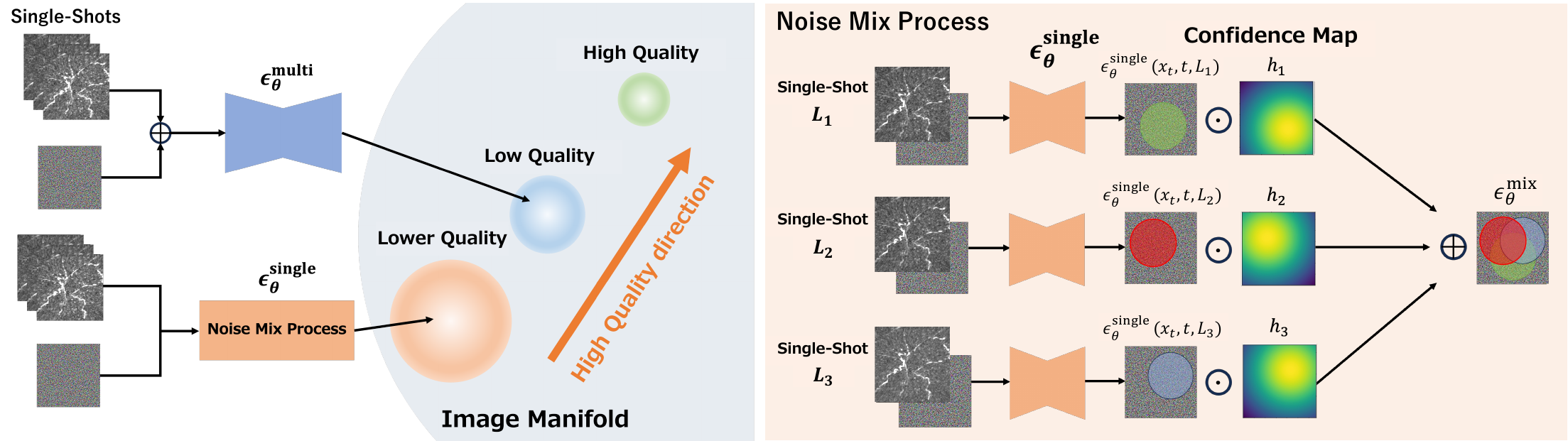}
    \vspace{2mm}
    \caption{Overview of proposed method, which consists guidance toward higher quality images and Noise Mix Process with photoacoustic imaging condition}
    % \vspace{1mm}
    \label{fig:proposed}
\end{figure*}

\section{Guidance-based diffusion model for improving photoacoustic image quality}
Given $M$ single-shot images ${L_{1}^i, \ldots, L_{M}^i}$ capturing the same location $i$, the proposed method estimates the high-quality image $H_i$, in which the quality of $H_i$ is comparable to that of an image generated by averaging over 20 images, and $M<<20$.
In addition, we also use the corresponding scattering light distribution maps $\{\bm{h}^i_1, \ldots, \bm{h}^i_M\}$ for this task.

To achieve this, we integrate a guidance mechanism into the backbone method, Denoising Diffusion Probabilistic Models (DDPM)~\cite{ho2020denoising}. Our approach utilizes the low-quality images $\{L_{1}^i, \ldots, L_{M}^i\}$ as conditions and denoises the random noise to generate the corresponding high-quality image in a reverse process. Within this reverse process, we introduce guidance defined by the vector from the noise estimated by a lower quality (a single-shot image) to that estimated by the $M$ single-shot images.

\subsection{Denoising Diffusion Probabilistic Models}
%This method uses Denoising Diffusion Probabilistic Models (DDPM)~\cite{} as a backbone model.
%we utilize a model based on the framework of Denoising Diffusion Probabilistic Models (DDPM). 
In this section, we explain the backbone method DDPM.
In the diffusion process of DDPM, noise is progressively added to an original image $\bm{x}_0$, which is a high-quality image $H$, from timestep $t=1$ to $T$ according to the following equation:
\begin{equation}
    \bm{x}_t = \sqrt{1-\beta_t}\bm{x}_{t-1} + \sqrt{\beta_t}\bm{\epsilon},
    \label{eq:q_ddpm}
\end{equation}
where $\beta_t$ indicates the intensity of the noise, and $\bm{\epsilon} \sim N(0,I)$ represents random noise. The reverse diffusion process is defined from $t=T$ to $1$ based on the following equation, utilizing a model $\bm{\epsilon_{\theta}}$ with parameters $\theta$:
\begin{equation}
    \bm{x}_{t-1} = \frac{1}{\sqrt{1-\beta_t}}(\bm{x}_t - \frac{\beta_t}{\sqrt{1-\alpha_t}}\bm{\epsilon}_\theta(\bm{x}_t,t)) + \sigma_t{\bm{z}},
    \label{eq:p_ddpm}
\end{equation}
where $\alpha_t=\prod^{t}_{s=1}(1-\beta_s)$, $\sigma_t=\sqrt{\beta_t}$, and $\bm{z}$ is random noise following $N(0,I)$. For image transformation with DDPM, a conditional model $\bm{\epsilon}_{\theta}(\bm{x}_t,t,c)$, where the input image $c$ serves as the condition, is used. DDPM achieves image generation by estimating noise with the model $\bm{\epsilon}_\theta$. Therefore, the model is trained to minimize the following loss function:
\begin{equation}
    Loss = \mathbb{E}_{t,\bm{x}_0,\bm{\epsilon}}[\parallel{\bm{\epsilon} - \bm{\epsilon}_\theta(\bm{x}_t,t,c)}\parallel].
    \label{eq:loss_ddpm}
\end{equation}
Based on this DDPM framework, our method inputs single-shot PA images as the condition and generates corresponding high-quality images. The condition is attached and provided at the network's input layer, i.e., single-shot images are concatenated to a noise image $\bm{x}_t$.
\subsection{Guidance toward higher quality}\label{Sec:Guidance}
In the generative process of diffusion models, guiding from lower (a single-shot) to low (few-shot images) quality can potentially result in higher quality outputs. In our approach, generation solely based on a single shot is deemed lower quality, while generation based on multiple shots is considered low quality. We offer guidance to produce higher-quality PA images.

Figure~\ref{fig:proposed} (Left) shows the overview of our method.
To introduce guidance, we prepare two models $\bm{\epsilon}^{\mathrm{single}}_{\theta}$ and $\bm{\epsilon}_{\theta}^{\mathrm{multi}}$, each trained with single-shot and multi-shots as conditions, respectively. 
Both models generate corresponding high-quality images based on the single-shot or few-shot images provided as conditions.

In the model $\bm{\epsilon}_{\theta}^{\mathrm{multi}}$, multiple-shot images are used as conditions. In contrast, in the model $\bm{\epsilon}^{\mathrm{single}}_{\theta}$, a single-shot image is used as a condition. This implies that the multi-shot model $\bm{\epsilon}_{\theta}^{\mathrm{multi}}$ can use richer information as conditions during the reverse diffusion process compared to the single-shot model $\bm{\epsilon}^{\mathrm{single}}_{\theta}$.
Therefore, $\bm{\epsilon}_{\theta}^{\mathrm{multi}}$ is capable of higher quality generation than $\bm{\epsilon}^{\mathrm{single}}_{\theta}$.
This is because, due to photoacoustic imaging, even single-shot images captured at the same location hold different structural information, and image averaging can reduce noises; thus, inputting more single-shot images results in higher accuracy. 

We guide towards further quality improvement by using the difference in outputs from models trained with single-shot and multi-shots. Specifically, in each timestep of the reverse diffusion process, instead of $\bm{\epsilon}_\theta(\bm{x}_t,t,c)$, we use $\tilde{\bm{\epsilon}}_\theta(\bm{x}_t,t,c)$ calculated based on the following:
\begin{equation}
    \tilde{\bm{\epsilon}}_\theta(\bm{x}_t,t,c) = (1+w)\bm{\epsilon}_{\theta}^{\mathrm{multi}}(\bm{x}_t,t,\{L_{1}^i, \ldots, L_{M}^i\}) - w \bm{\epsilon}_\theta^{\mathrm{mix}}(\bm{x}_t,t,c) ,
    \label{eq:cond_guidace}
\end{equation}
where ${\bm{\epsilon}}_\theta^{\mathrm{mix}}(x_t,t,c)$ is the output obtained through the Noise Mix Process using $\bm{\epsilon}_{\theta}^{\mathrm{single}}$, described in the next section. Additionally, $w$ is a hyperparameter indicating the strength of guidance. Through equation~(\ref{eq:cond_guidace}), we can expand the difference between the consistent outputs of $\bm{\epsilon}_{\theta}^{\mathrm{single}}$ and $\bm{\epsilon}_{\theta}^{\mathrm{multi}}$ by $w$ and guide towards generating higher quality photoacoustic images.

%In the model $\bm{\epsilon}_{\theta^{\mathrm{multi}}}$, a rich information is inputted compared to $\bm{\epsilon}_{\theta^{\mathrm{single}}}$ since 

%In our problem setup, we have several single-shot images corresponding to $H_i$. If we randomly select one of the single-shot images, as discussed in Section 3, 

%The overview of our guidance method is shown in Figure~\ref{fig:proposed} left. First, we prepare two models $\bm{\epsilon}{\theta}^{\mathrm{single}}$ and $\bm{\epsilon}{\theta}^{\mathrm{multi}}$, each trained with single-shot and multi-shots as conditions, respectively. Both models generate corresponding high-quality images based on the one-shot images provided as conditions. However, given that the number of one-shot images as conditions differs, the model $\bm{\epsilon}_{\theta}^{\mathrm{multi}}$, which has more conditions, is considered capable of higher quality generation. This is because, due to the nature of photoacoustic imaging, even one-shot images captured at the same location hold different structural information; thus, inputting more one-shot images results in higher accuracy. Using the difference in outputs from models trained with single-shot and multi-shots, we guide towards further quality improvement.Specifically, in each timestep of the reverse diffusion process, instead of $\bm{\epsilon}_\theta(\bm{x}t,t,c)$, we use $\tilde{\epsilon\theta}(\bm{x}_t,t,c)$ calculated based on the following:

\subsection{Noise mix process of single-shot image-based models} 
Our problem setup includes several single-shot images corresponding to $H_i$, each exhibiting different scattering light distribution maps as illustrated in Figure~\ref{fig:pa}. When we randomly select one of these single-shot images, certain areas may exhibit high intensity, which has high reliability, depending on the light distribution map. In contrast, others show low intensity. This variability could potentially worsen the effectiveness of the guidance.

To address this issue, we introduce a noise-mixing technique to generate a higher-quality image by incorporating clear parts from each single-shot image. The noise mixing technique involves interpolating between two images~\cite{song2021denoising} and controlling image generation from text~\cite{shirakawa2024noisecollage}. Inspired by these methods, we propose a weighted noise mixing process to leverage the unique properties of image averaging.

The noise mix process combines the estimated noise images 
$\bm{\epsilon}_{\theta}^{\mathrm{single}}(L_1), \ldots,\bm{\epsilon}_{\theta}^{\mathrm{single}}(L_M)$ using their corresponding light distribution maps $\{\bm{h}^i_1, \ldots, \bm{h}^i_M\}$.
As mentioned above, in PA imaging, the reliability of the acquired signal changes depending on the distance from the light irradiation position, and these positions vary in each one-shot image. Therefore, a light distribution map $\bm{h}^i_m$ can be considered as a confidence map. Based on the assumption that the estimated noise in high-confidence areas is more reliable than that in low-confidence areas, our method combines the estimated noise images through a weighted average, where the weights are determined by the confidence maps as follows:
%Therefore, a confidence map is created for each one-shot image based on the light irradiation position. 
%First, a heatmap based on a Gaussian distribution centered around the position of light irradiation is created (with values ranging from 0.2 to 1.0). 
%From this heatmap, the overlapping parts of one-shot images during image averaging are used as the confidence map for each one-shot image.
%In the noise mix process, $\bm{\epsilon}_{\theta}^{\mathrm{single}}$ is provided with each one-shot image individually, and $\bar{\bm{\epsilon}}\theta$ is calculated based on the following equation from the outputs corresponding to each one-shot image and the aforementioned confidence maps:
\begin{equation}
   \bm{\epsilon}_\theta^{\mathrm{mix}}(\bm{x}_t,t,c) = \frac{1}{M}\sum^M_{m=1}{\bm{h}_m^i \odot \bm{\epsilon}_{\theta}^{\mathrm{single}}(\bm{x}_t,t,L_m^i)}.
    \label{eq:noise mix}
\end{equation}
where $\bm{x}_t$ is the noise image at $t$ step in diffusion process, $\bm{\epsilon}_{\theta}^{\mathrm{single}}(\bm{x}_t,t,L_m^i)$ represents the output of $\bm{\epsilon}_{\theta}^{\mathrm{single}}$ corresponding to each single-shot image $L_m^i$. $\bm{h}_m^i$ is the confidence map for the single-shot image $L_m^i$. 

Mixing the noise estimated from individual single-shot images using equation~\ref{eq:noise mix} allows for the integration of their respective structural information. Furthermore, considering the confidence maps based on the light irradiation positions of each single-shot image enables more accurate completion of the foreground parts.

\section{Experiments}
\noindent\textbf{Dataset.} We used real PA images for evaluation. The images were taken of the lower limbs of two subjects, who were instructed to remain still during data collection.
For the training, validation, and test data, we prepared paired images of low and high-quality images $\{\{L_{1}^i, \ldots, L_{M}^i\}, H^i\}_{i=1}^N$ with their light distribution maps $\{\bm{h}^i_1, \ldots, \bm{h}^i_M\}$.
The number of single-shot images $M$ was 3.
The training, validation, and test data contain 3988, 1328, and 907 pairs, respectively. For test data, we used different subjects' PA images with training and validation data.
%consist of 3988 pairs, and the test data consists of 907 pairs derived from different subjects' PA images.
%real PA images captured from one-shot images. 
%We created a dataset where the input images are single-shot images, and the corresponding high-quality images enhanced by image averaging served as the teacher images. Since the goal of this study is to achieve high-quality results from a few one-shot images, we set the number of one-shot images used as inputs to three. The training data consisted of 3988 pairs, and the test data consisted of 907 pairs, both derived from PA images of different subjects.

\noindent\textbf{Implimentation Details.} We used a U-Net-like network~\cite{ho2020denoising} as the diffusion model $\bm{\epsilon}_{\theta}^{single}$, $\bm{\epsilon}_{\theta}^{multi}$. The U-Net-based diffusion model has the residual layer and self-attention to improve the model's representation performance. The training was conducted over 300,000 iterations with a batch size of 16. The optimization algorithm used was Adam, with a learning rate of $1.0\times10^{-4}$. The total number of timesteps $T$ in sampling and the noise scheduler were the same as those in \cite{ho2020denoising}, with $T=1000$, $\beta_t = 10^{-4}$, and $\beta_T = 0.02$.

There is a known issue that excessively large guidance scales $w$ can degrade the quality of image generation\cite{saharia2022photorealistic}. In the generative process of diffusion models, semantic information is formed in the early stages, while finer image details are developed towards the end \cite{choi2022perception, diffusion2023}. Thus, especially towards the end, using a large guidance scale $w$ can lead to deviations from the training data distribution of the diffusion model, resulting in degraded generation quality. Therefore, our method sets an interval $[T,t_{guide}]$ where guidance is used with a predefined $w$, and $[t_{guide},0]$ where $w=0$. This $t_{guide}$ was tuned using validation data.

\noindent\textbf{Comparative Methods.} As the most naive approach, we used image averaging of three input images as the Baseline. For comparison, we employed a CNN-based method with U-Net \cite{Ronneberger2015} and DnCNN \cite{zhang2017beyond} for image transfer. As the state-of-the-art methods, a Transformer-based method with Restormer \cite{zamir2022restormer}, and a standard diffusion model, DDPM \cite{ho2020denoising} were evaluated. Here, DDPM refers to $\bm{\epsilon}_{\theta}^{\mathrm{multi}}$, given three single-shot images as conditions and performing conditional generation based solely on equation~(\ref{eq:p_ddpm}).
Each deep neural network-based method was trained with the set of three one-shot images as input.

\noindent\textbf{Evaluation Metrics.} To compare the accuracy of the outputs from each method, Peak Signal to Noise Ratio(PSNR) and Structural Similarity Index Measure(SSIM) between the estimated high-quality image by each method and the ground truth were used, which have been widely used in image denoising tasks. The higher their value, the higher the similarity.
\begin{figure*}[t]
    \centering
    \includegraphics[scale=0.35]{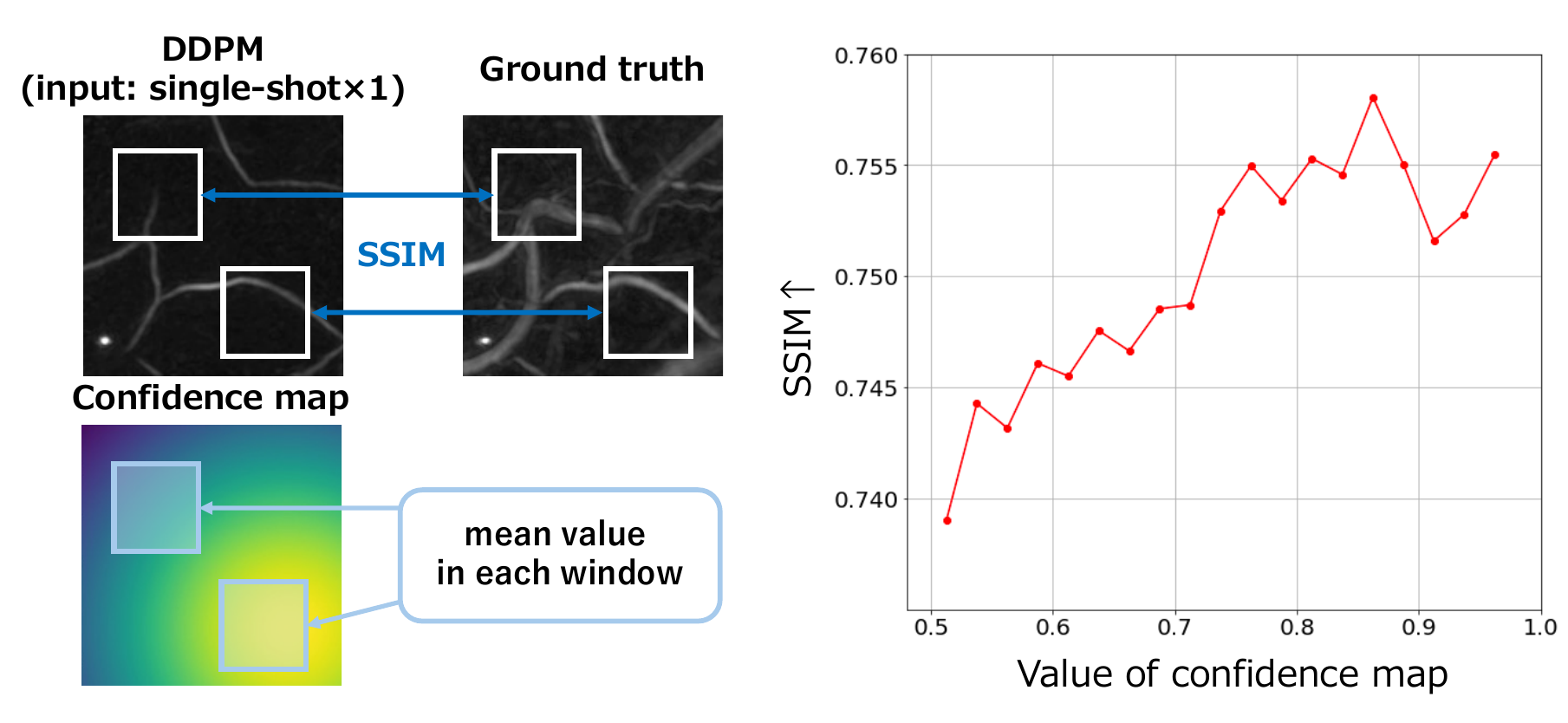}
    \vspace{2mm}
    \caption{Preliminary experiments with confidence maps, {\bf Left}: overview of evaluation method, {\bf Right}: correlation of SSIM and confidence maps}
    \label{fig:pre-exp1}
    \vspace{-2mm}
\end{figure*}
\subsection{Correlation with Confidence Maps}
% First, we conducted preliminary experiments to quantitatively verify how the clarity of foreground parts (blood vessels) in each shot image varies with the distance from the light irradiation position.

% Figure~\ref{fig:pre-exp1} illustrates the relationship between SSIM values and the corresponding confidence map values within a local area (window size 50×50) of DDPM outputs conditioned on a single one-shot image and the ground truth. We generated a scatter plot by plotting the mean values of SSIM and confidence map values across all test data, segmented into 20 bins along the scatter plot's horizontal axis to calculate and plot their average values.

% As shown in Figure~\ref{fig:pre-exp1}, SSIM increases with the confidence map values within the window, indicating that higher values enable more accurate predictions, while lower values present challenges. Hence, using multiple one-shot images as inputs with their respective confidence maps can enhance prediction accuracy.

First, we conducted preliminary experiments to quantitatively verify that the clarity of the foreground parts (blood vessels) changes depending on the distance from the light irradiation position in each shot image. 

%Figure.\ref{fig:pre-exp1} shows the relationship between the SSIM values in a local area (window size 50×50) of the output from DDPM with a single-shot image as the condition and the ground truth in the test data and the values of the corresponding confidence map in the same area.
Figure~\ref{fig:pre-exp1} (Right) illustrates the relationship between the mean SSIM values of an image estimated by DDPM with a single-shot image in a local area (window size 50×50) and the mean confidence values of the light distribution map in the same area.
To compute the average of the SSIM values, we divided the confidence values into 20 equal ranges, and calculated the means of SSIMs for all locations in all test data within each range, as plotted in Figure~\ref{fig:pre-exp1} (Right).
As a result, the SSIM increased as the confidence values increased.
%Initially, we created a scatter plot by plotting the average values of SSIM and the confidence map within each randomly set window across all test data. Subsequently, this scatter plot's horizontal axis was divided into 20 bins, and the average values of each point within those intervals were calculated and plotted in the figure.

Figure~\ref{fig:pre-exp1} (Left) shows example images of the estimated high-quality image by DDPM, its ground truth, and the corresponding confidence map. The vessel clearly appears in the region with high confidence values (right-bottom area in the map). In contrast, the vessel is unclear in the region with low confidence values (left-top).

These results suggest that considering the corresponding confidence maps for each can enable more accurate predictions when using multiple single-shot images as inputs.
\begin{table}[t]
    % \def\@captype{table}
    %   \makeatother
        \begin{minipage}[t]{0.5\textwidth}
          \vspace{-13mm}
          \centering
          % \vspace{7mm}
          % \scalebox{1}{
          \begin{tabular}{llcc}
            \hline
            \multicolumn{2}{c}{Method}  & PSNR↑  &  SSIM↑   \\
            \hline \hline
            \multicolumn{2}{l}{Baseline}  & 20.63  & 0.3396 \\
            \multicolumn{2}{l}{U-Net\cite{Ronneberger2015}}  & 30.10  & 0.5055 \\
            \multicolumn{2}{l}{DnCNN\cite{zhang2017beyond}} & 30.14  & 0.5210 \\
            \multicolumn{2}{l}{Restormer\cite{zamir2022restormer}}  & 30.48  & 0.5247 \\
            \multicolumn{2}{l}{\bf{Ours}} & \bf{30.63}  & \bf{0.5468} \\

            \hline
          \end{tabular}
          \vspace{5mm}
          \caption{Comparative experiments with \\ previous methods}
          \label{table:t1}
          % }
        \end{minipage}
        \begin{minipage}[t]{0.4\textwidth}
          % \vspace{-3mm}
          \centering
          % \scalebox{1}{
          \begin{tabular}{llcc}
            \hline
            \multicolumn{2}{c}{Method}  & PSNR↑  &  SSIM↑   \\
            \hline \hline
            \multirow{2}{*}{$w=5$} & \bf{Ours}  &  30.27   & 0.5132 \\
                                   & \bf{w/o $h$}  &  30.26   & 0.5144 \\
            % \quad \quad \,\,\, w/o $h$  & 30.06   & 0.4883 \\
            \hline
            \multirow{2}{*}{$w=10$} & \bf{Ours}  & \bf{30.63}   & \bf{0.5468} \\
                                   & \bf{w/o $h$}  & 30.61   & 0.5456 \\
            % \quad \quad \,\,\, $w=10$  & \bf{30.60}   & \bf{0.5396} \\
            % \quad \quad \,\,\, w/o $h$  & 30.59   & 0.5381 \\
            \hline
            \multirow{2}{*}{$w=20$} & \bf{Ours}  & 30.29   & 0.5313 \\
                                   & \bf{w/o $h$}  & 30.25   & 0.5301 \\
            % \quad \quad \,\,\, $w=20$  & 30.11   & 0.5338 \\
            % \quad \quad \,\,\, w/o $h$  & 29.88   & 0.5347 \\
            \hline
            \multirow{2}{*}{$w=30$} & \bf{Ours}  & 27.94   & 0.4611 \\
                                   & \bf{w/o $h$}  & 27.70   & 0.4573 \\
            % \quad \quad \,\,\, $w=30$  & 26.83   & 0.4476 \\
            % \quad \quad \,\,\, w/o $h$  & 26.69   & 0.4440 \\

            \hline
          \end{tabular}
          \vspace{4mm}
          \caption{Ablation experiments with different guidance scales}
          \label{table:t2}
          % }
        \end{minipage}
\end{table}
\begin{figure}[t]
  \def\@captype{table}
  \begin{minipage}[c]{0.4\textwidth}
      \centering
      \small
      % \vspace{-5mm}
      \vspace{5mm}
      \begin{tabular}{llcc}
            \hline
            \multicolumn{2}{c}{Method}  & PSNR↑  &  SSIM↑   \\
            \hline \hline 		
            % \multicolumn{2}{l}{DDPM(one-shot×1)}  & 28.88  & 0.3889 \\
            \multicolumn{2}{l}{DDPM\cite{ho2020denoising}}  & 29.39  & 0.4159 \\
            \multicolumn{2}{l}{Classifier-Free Guidance\cite{ho2021classifierfree}}  & 29.38  & 0.5315 \\
            \multicolumn{2}{l}{Ours w/o Noise Mix Process}  & 30.38  & 0.5447 \\
            \multicolumn{2}{l}{Ours} & 30.63 & 0.5468 \\

            \hline
        \end{tabular}
        \vspace{5mm}
        \tblcaption{Ablation experiments with guidance condition}
        \label{table:t3}
  \end{minipage}
  % \hfill
  \begin{minipage}[c]{0.7\textwidth}
    \centering
    \vspace{-2mm}
    \includegraphics[scale=0.38]{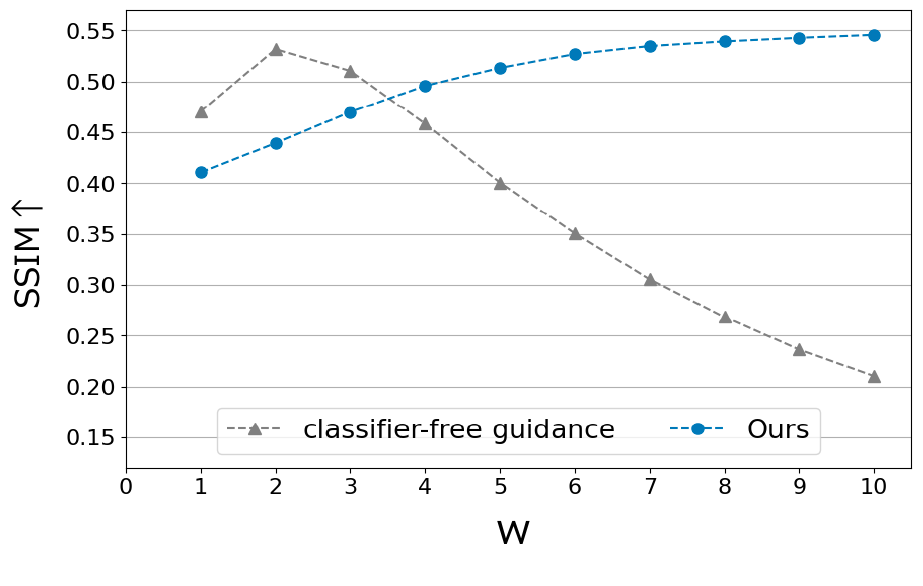}
    \vspace{1mm}
    \caption{guidance scale-wise SSIM}
    \vspace{-7mm}
    \label{fig:exp4}
  \end{minipage}
\end{figure}
\subsection{Quantitative evaluation}
\noindent\textbf{Comparative study.}
Table~\ref{table:t1} presents the average performances (PSNR and SSIM) based on evaluation metrics for each method's output results on the test data. Deep learning-based methods significantly improved image quality compared to the simple image averaging of the Baseline. It is confirmed that the proposed method shows the best results in both PSNR and SSIM, particularly in SSIM, which is a metric that assumes the similarity of image structures contributes to human perception of image quality degradation.

\noindent\textbf{Hyper-parameter sensitivity.}
Table~\ref{table:t2} shows the results of experiments conducted by varying the guidance scale $w$ without using the confidence map $h$. First, comparing the variations in $w$ within our method, a decline in accuracy is observed at $w=30$. It is confirmed that excessively increasing the guidance scale can deteriorate the generative results. 

Next, w/o $h$ in Table~\ref{table:t1} refers to averaging the outputs based on each single-shot image in our method's noise mix process without utilizing the corresponding confidence map $h$. 
The results show that using the confidence map $h$ achieves higher PSNR values for each $w$, and although SSIM values are higher without the confidence map at $w=5$, using the confidence map at $w=10$ consistently shows the best results. Here, the best $w=10$ can be selected using validation data.
This suggests that considering the confidence of each single-shot image contributes to improved accuracy.

\noindent\textbf{Ablatoin study.}
Table~\ref{table:t3} also presents the results of the ablation study of our method. Here, variations in the guidance methods were tested: DDPM was executed without guidance, using three single-shot images as conditions, classifier-free guidance was based on unconditional noise, and w/o Noise Mix Process directly used the output from $\bm{\epsilon}_{\theta}^{\mathrm{single}}$ as the basis for guidance. The simplest DDPM output was the least accurate, and accuracy improved with each guidance method. Notably, the proposed method of introducing the noise mix process based on imaging conditions achieved the highest performance. 

Additionally, in classifier-free guidance and the proposed method, the results of varying the guidance scale from 1 to 10 are shown in Figure~\ref{fig:exp4}. In classifier-free guidance, accuracy declined starting from $w=3$, whereas in the proposed method, accuracy consistently improved from $w=1$ to $10$. It is considered that the guidance method of the proposed approach, using correlated low-quality and high-quality outputs, allows for guidance toward higher-quality image generation.

\begin{figure*}[t]
    \begin{tabular}{c}
    \centering
        \begin{minipage}{0.6\textwidth}
            \vspace{1mm}
            \centering
            \includegraphics[width=1.0\linewidth]{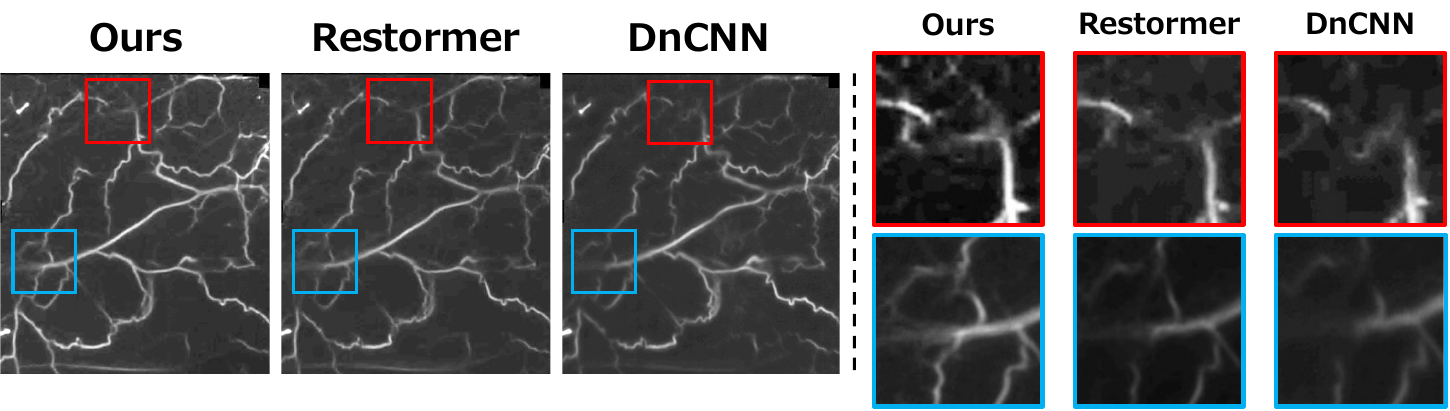}
            \vspace{3mm}
            \caption{{\bf Left:} Wide-view PA image obtained by ours and other methods, {\bf Right:} Enlarged images.}
            \label{fig:show1}
        \end{minipage}%
        \begin{minipage}{0.01\textwidth}
            \hfill
        \end{minipage}
        \begin{minipage}{0.35\textwidth}
            \centering
            \includegraphics[width=1.0\linewidth]{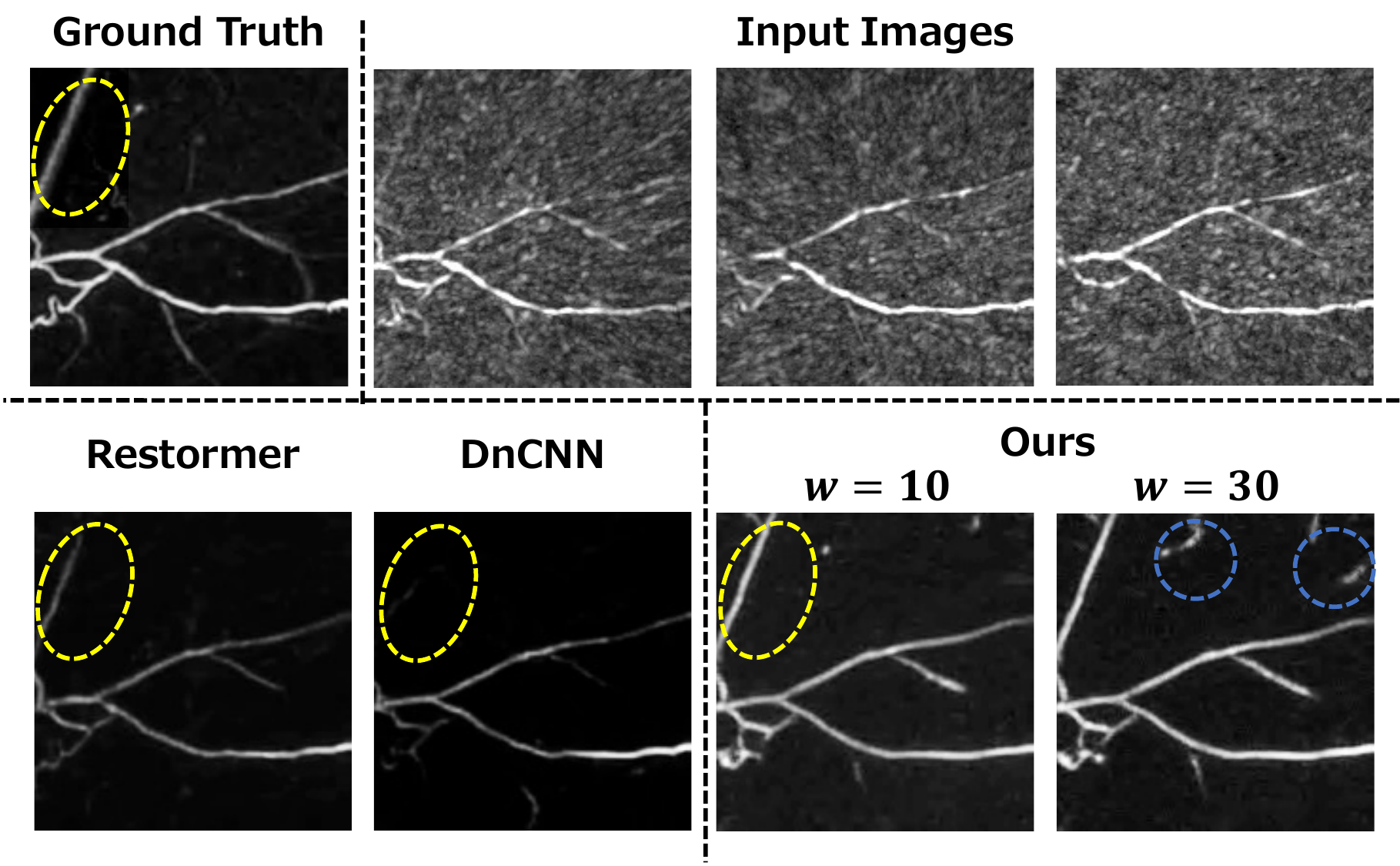}
            \vspace{1mm}
            \caption{Example of complementation for corruption}
            \label{fig:show2}
        \end{minipage}
    \end{tabular}
    \vspace{-3mm}
\end{figure*}

\subsection{Qualitative evaluation}
Figure~\ref{fig:show1} shows example wide-view images of the results for our method, Restormer \cite{zamir2022restormer}, and DnCNN \cite{zhang2017beyond}. The small images on the right in Figure~\ref{fig:show1} are enlarged areas of the red and blue boxes in the wide-view images.
In these results, some blood vessels were missing in the comparison methods. In contrast, the proposed method made the blood vessels clearer, as shown in the enlarged images.

Figure~\ref{fig:show2} shows examples of ground truth, single-shot images(input images), and output results of each method. In the results, noise in the background parts contained in the single-shot images was removed using all methods. 
Focusing on the parts highlighted in yellow, the comparative methods do not sufficiently complete the missing blood vessels, whereas our method successfully completed these missing vessels.

The qualitative evaluation was conducted for different guidance scales $w$. Regarding the output results at $w=30$ in our method, as indicated by the blue frames, blood vessels are complemented in areas where, according to ground truth, they should not exist. This suggests that excessively increasing the guidance scale can lead to over-detection of blood vessels, adversely affecting the generation results. In contrast, the suitable scale of $w$, which is automatically selected using validation data, shows a good result.

\section{Discussion and Conclusion}
In this study, we proposed a method for enhancing the quality of photoacoustic(PA) images from a small number of single-shot images to reduce imaging costs in PA imaging.
%\par
%In PA imaging, acquiring many single-shot images and performing image averaging can produce clear PA images. However, capturing many one-shot images is time-consuming and costly. Therefore, by enhancing the quality of PA images using a few one-shot images, it is possible to reduce imaging costs.
\par
We introduce a guidance approach that considers the reliability of the signals obtained at the time of imaging, and it improves the reverse diffusion process in diffusion models. The structural information from multiple single-shot images is effectively reflected in the generation results by defining and utilizing a confidence map based on the light irradiation position during imaging. Moreover, we introduced a guidance technique for diffusion models that leads to higher-quality generation results. Experiments using real data from PA images demonstrated the effectiveness of our method compared to traditional techniques.
\par
As with most applications with diffusion models, our method is limited primarily by slow inference times. However, our main goal is to achieve reconstruction with a few one-shot images, which, if successful, could reduce the long scan times typically required in PA imaging. Also, our method can explore acceleration techniques as in \cite{rombach2022high, Song2023ConsistencyM, song2024improved, liu2023flow}. 
\par
We believe that our guidance technique, which utilizes vectors from noise estimated in lower-quality images to that in higher-quality images for diffusion models, can be applied to any denoising tasks, not just PA imaging.

\vspace{0.5\baselineskip}
%============================================================
\section*{Acknowledgment}
%============================================================
This work was supported by SIP-JPJ012425, AMED JP19he2302002, JSPS KAKENHI Grant JP23K18509 and JP24KJ1805.

\bibliography{egbib}
\end{document}